\documentclass{article}

\usepackage{microtype}
\usepackage{graphicx}
\usepackage{subcaption}
\usepackage{booktabs} 

\usepackage{hyperref}

\usepackage[preprint]{icml2026}

\usepackage{amsmath}
\usepackage{amssymb}
\usepackage{mathtools}
\usepackage{amsthm}

\usepackage[capitalize,noabbrev]{cleveref}

\usepackage{booktabs}
\usepackage{multirow}
\usepackage{array}
\usepackage{booktabs}   
\usepackage{tabularx}   
\usepackage[table]{xcolor} 

\theoremstyle{plain}

\theoremstyle{definition}

\theoremstyle{remark}

\usepackage[textsize=tiny]{todonotes}

\icmltitlerunning{Diffusion LMs Can Approximate Optimal Infilling Lengths Implicitly}

\begin{document}

\twocolumn[
  \icmltitle{Diffusion LMs Can Approximate Optimal Infilling Lengths Implicitly}



  \icmlsetsymbol{equal}{*}

  \begin{icmlauthorlist}
    \icmlauthor{Hengchang Liu}{equal,ruc}
    \icmlauthor{Zhao Yang}{equal,ruc}
    \icmlauthor{Bing Su}{ruc}
  \end{icmlauthorlist}

  \icmlaffiliation{ruc}{Gaoling School of Artificial Intelligence, Renmin University of China}

  \icmlcorrespondingauthor{Bing Su}{bingsu@ruc.edu.cn}

  \icmlkeywords{Diffusion Language Models, Generative Infilling, Adaptive Length Control, Training-Free Generation}

  \vskip 0.3in
]



\printAffiliationsAndNotice{\icmlEqualContribution}

\begin{abstract}
  Diffusion language models (DLMs) provide a bidirectional generation framework naturally suited for infilling, yet their performance is constrained by the pre-specified infilling length. In this paper, we reveal that DLMs possess an inherent ability to discover the correct infilling length. We identify two key statistical phenomena in the first-step denoising confidence: a local \textit{Oracle Peak} that emerges near the ground-truth length and a systematic \textit{Length Bias} that often obscures this signal. By leveraging this signal and calibrating the bias, our training-free method \textbf{CAL} (\textbf{C}alibrated \textbf{A}daptive \textbf{L}ength) enables DLMs to approximate the optimal length through an efficient search before formal decoding. Empirical evaluations demonstrate that CAL improves Pass@1 by up to 47.7\% over fixed-length baselines and 40.5\% over chat-based adaptive methods in code infilling, while boosting BLEU-2 and ROUGE-L by up to 8.5\% and 9.9\% in text infilling. These results demonstrate that CAL paves the way for robust DLM infilling without requiring any specialized training. Code is available at \url{https://github.com/NiuHechang/Calibrated_Adaptive_Length}.
\end{abstract}

\section{Introduction}
\label{sec:introduction}

Recently, Diffusion Language Models (DLMs) \cite{gong2024scaling, llada, ye2025dream} have demonstrated significant potential in natural language processing. Unlike traditional Autoregressive (AR) models that rely on token-by-token causal generation, diffusion models generate sequences via iterative denoising over the entire sequence \cite{ddpm, discrete_diffusion}. While mainstream DLMs predominantly focus on chat tasks akin to AR models, this usage underutilizes their unique potential. With their natural bidirectional context modeling, diffusion models are inherently suited for infilling tasks \cite{lewis2020bart, enable_infilling} that require satisfying constraints from both the prefix and suffix. This capability is particularly critical for applications such as code completion \cite{humaneval_infilling, fried2022incoder}, text infilling \cite{text_infilling, lewis2020bart}, and controllable generation \cite{zhou2023controlled, controllable_generation}, where the generated segment must syntactically and logically bridge the surrounding context.

\begin{figure*}[ht]
  \vskip 0.1in
  \begin{center}
    \centerline{\includegraphics[width=2\columnwidth]{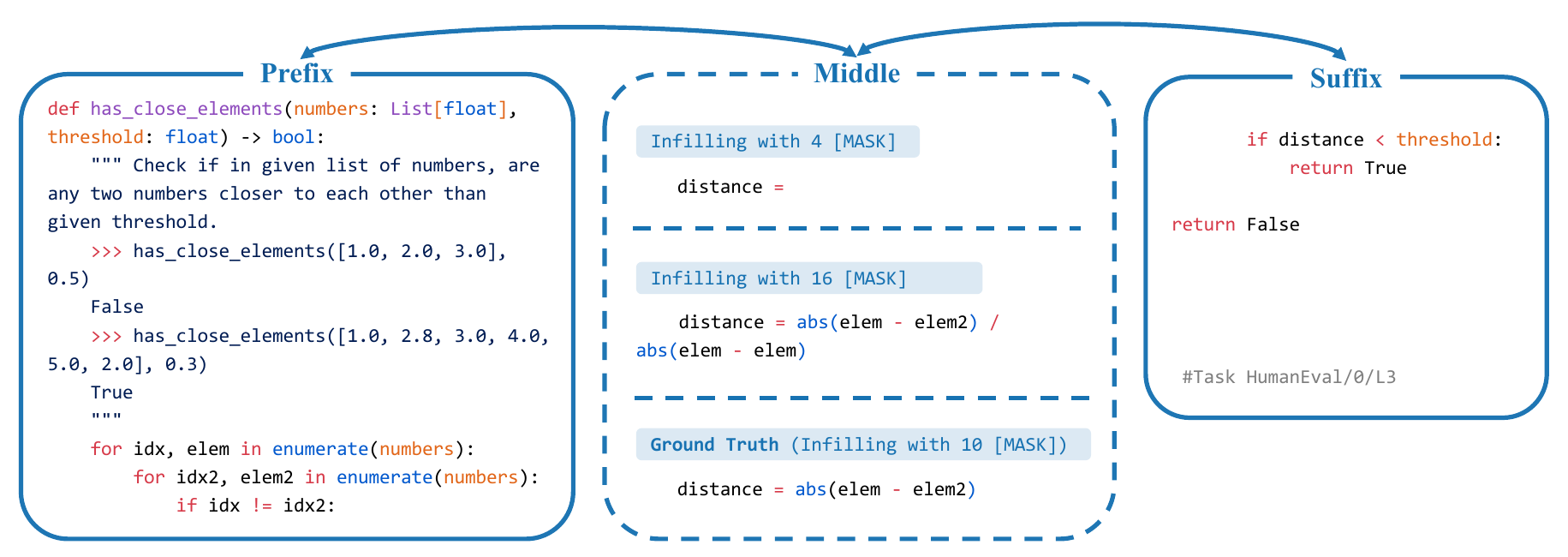}}
        \caption{\textbf{Infilling sensitivity to a fixed mask length.} Evaluated on HumanEval-Infilling \cite{humaneval_infilling} using LLaDA-8B-Instruct. An underestimated length ($L=4$) truncates the completion, whereas an overestimated length ($L=16$) introduces redundant tokens and may yield invalid logic (e.g., division by zero) to fill the remaining masked positions. The correct solution is only recovered when the length matches the ground truth ($L=10$).}
    \label{fig:infilling_example}
  \end{center}
  \vskip -0.2in
\end{figure*}

Despite this theoretical promise, the actual performance of DLMs in infilling tasks often falls short of expectations (see \Cref{sec:analysis}). We find that this performance gap primarily stems from the lack of adaptive length adjustment in DLMs as quantified in \Cref{tab:pass1_analysis}. In chat-based tasks, DLMs can generate a specific \texttt{<eos>} token to terminate generation \cite{llada, gong2024scaling, ye2025dream}. In contrast, for infilling tasks, these models lack an intrinsic stopping mechanism and thus rely on a fixed, pre-specified mask length \cite{diffucoder, dreamcoder}. Without a dynamic mechanism, the generation quality of DLMs is highly sensitive to the pre-specified mask length. As shown in \Cref{fig:infilling_example}, an underestimated infilling length truncates the completion, whereas an overestimated length causes the model to generate redundant or incorrect tokens to fill the remaining masked positions. This length constraint significantly limits DLM performance on infilling tasks.

\begin{figure*}[t]
  \centering
  \begin{subfigure}[b]{0.49\textwidth}
    \centering
    \includegraphics[width=\linewidth]{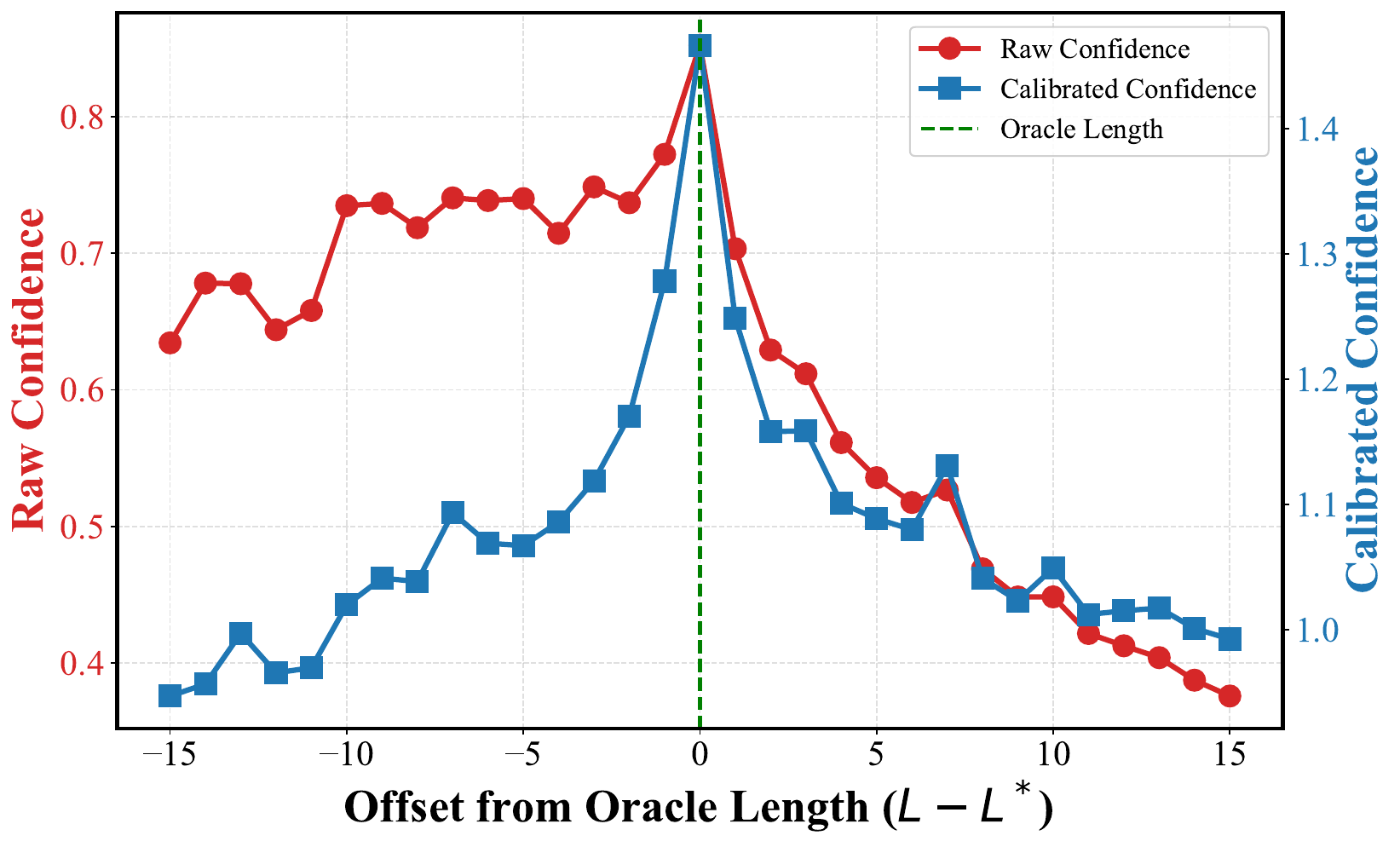}
    \caption{\textbf{Oracle Peak Phenomenon.} Confidence peaks at $L - L^* = 0$, indicating maximal semantic consistency when the generation window matches the ground-truth length. The red line represents raw confidence, while the blue line shows confidence after bias calibration.}
    \label{fig:oracle_peak}
  \end{subfigure}
  \hfill 
  \begin{subfigure}[b]{0.45\textwidth}
    \centering
    \includegraphics[width=\linewidth]{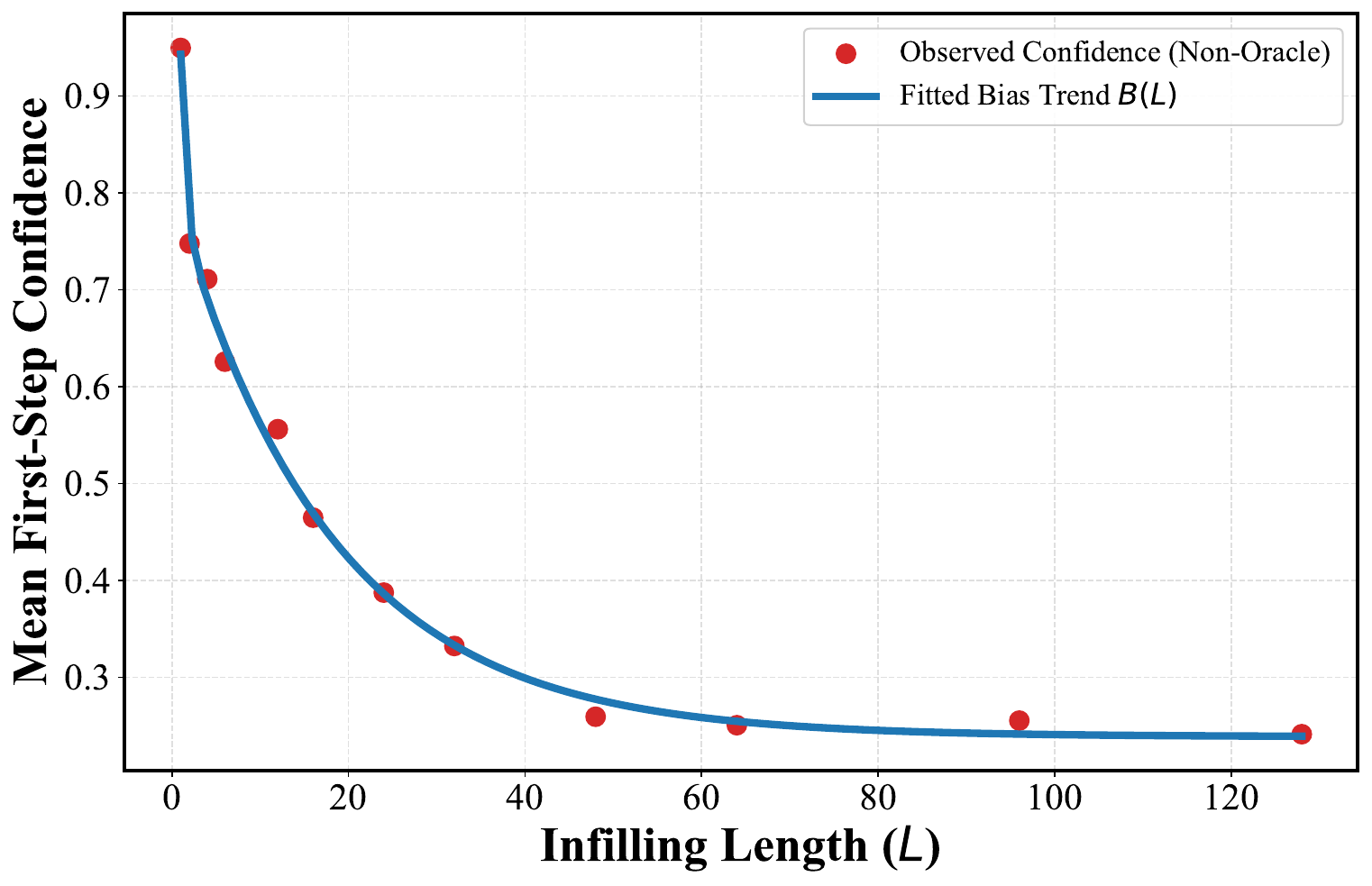}
    \caption{\textbf{Length Bias Fitting.} The systematic decay of first-step confidence as length increases. Red dots represent empirical $\Phi(L)$ samples computed from non-Oracle lengths (excluding $L \in [L^* - 4, L^* + 4]$); the blue curve shows the fitted double-exponential bias function $B(L)$.}
    \label{fig:length_bias}
  \end{subfigure}
 \caption{\textbf{Statistical Analysis of First-Step Denoising Confidence $\Phi(L)$.} (a) A distinct confidence peak emerges when aligned with the ground-truth length. (b) Without length information, confidence exhibits systematic decay, necessitating calibration. Statistics are derived from 100 random HumanEval-Infilling \cite{humaneval_infilling} tasks using LLaDA-8B-Base.}
  \label{fig:confidence_analysis}
\end{figure*}

To overcome this limitation without retraining, we investigate whether DLMs implicitly encode infilling length information within their generation process. We analyze the model's inherent confidence during first-step denoising from a fully masked state. By evaluating the average confidence across different infilling lengths, we observe two key statistical phenomena (see \Cref{fig:confidence_analysis}): (1) \textit{Oracle Peak}: under the same context, the average first-step confidence exhibits a local maximum when the mask length is close to the reference answer length; (2) \textit{Length Bias}: the average confidence decreases rapidly and then gradually stabilizes as the infilling length increases.

Mechanistically, the Oracle Peak suggests that first-step confidence encodes a notion of semantic completeness: when the infilling length matches the ground-truth span, the model assigns higher certainty to its initial token predictions (see \Cref{fig:oracle_peak}). In contrast, Length Bias reflects a systematic decline of average confidence as the number of masked positions increases in a fully masked state (see \Cref{fig:length_bias}): longer masked regions weaken contextual constraints and introduce more low-certainty positions, which lowers the average token-level confidence. By calibrating the first-step confidence with the estimated bias, the length-dependent trend is largely removed, making the Oracle Peak more salient. Overall, these observations suggest that calibrated first-step confidence can serve as a practical signal for infilling length discovery, enabling adaptive length control without additional training.

Based on these findings, we propose \textbf{CAL} (\textbf{C}alibrated \textbf{A}daptive \textbf{L}ength), a training-free method for infilling length discovery. CAL introduces a length probing stage prior to formal decoding (see \Cref{fig:method} and \Cref{alg:length_discovery}). Starting from an initial length, we evaluate the calibrated first-step confidence across a set of candidate lengths and perform a bidirectional hill-climbing search to approximate the optimal length. Since each probe requires only a single forward pass to compute the first-step logits, the additional inference overhead is modest.

Empirical evaluations on both code and text infilling benchmarks (see \Cref{tab:code_infilling,tab:text_infilling}) demonstrate that CAL consistently outperforms baselines across multiple DLMs. Specifically, it improves Pass@1 by up to 47.7\% over fixed-length baselines and 40.5\% over chat-based adaptive methods in code infilling, while boosting BLEU-2 and ROUGE-L by up to 8.5\% and 9.9\% in text infilling.  Our method decouples infilling performance from the rigid requirement of a pre-specified mask length, enabling more robust generation under bidirectional constraints. Our main contributions are: (1) revealing and modeling the \textit{Oracle Peak} and \textit{Length Bias} phenomena in first-step denoising for infilling; (2) proposing a training-free strategy for adaptive length probing via calibrated confidence; and (3) demonstrating the effectiveness of our method through extensive experiments.
\section{Preliminaries}
\label{sec:preliminaries}

\textbf{Diffusion Language Models.} Diffusion language models \cite{discrete_diffusion, hoogeboom2021argmax} generate sequences via iterative denoising with bidirectional context. Let $\mathbf{x}^0=[x^0_1,\dots,x^0_N]$ denote a clean sequence of length $N$. A forward noising process progressively corrupts $\mathbf{x}^0$ into $\mathbf{x}^t$ over $T$ steps (e.g., by random masking \cite{sahoo2024simple, shi2024simplified}), yielding a heavily corrupted state $\mathbf{x}^T$. A DLM defines a reverse generative process parameterized by $\theta$ that reconstructs $\mathbf{x}^0$ from $\mathbf{x}^T$:
\begin{equation}
p_\theta(\mathbf{x}^{0:T}) = p(\mathbf{x}^T)\prod_{t=1}^{T} p_\theta(\mathbf{x}^{t-1}\mid \mathbf{x}^t).
\end{equation}
In practice, many modern DLMs (e.g., LLaDA \cite{llada}) are trained to predict the clean tokens directly from a noisy state, i.e., to model $p_\theta(\mathbf{x}^0\mid \mathbf{x}^t)$. The resulting token distributions provide the confidence signal used throughout our analysis and length discovery method.

\begin{table*}[t]
 \caption{\textbf{Oracle length significantly improves DLM infilling performance.} Comparison of Pass@1 scores on HumanEval-Infilling \cite{humaneval_infilling} (Single-Line) between fixed-length settings and the Oracle length setting. The Oracle Length is determined by encoding the ground-truth answer with each model's tokenizer. ``Avg. Len'' indicates the average Oracle length across all test cases.}
  \label{tab:pass1_analysis}
  \begin{center}
    \begin{small}
      \begin{sc}
        \setlength{\tabcolsep}{10pt} 
        \begin{tabular}{l cccc c | cc}
          \toprule
          \multirow{2}{*}{Model} & \multicolumn{4}{c}{Fixed Length (Pass@1)} & \multirow{2}{*}{Avg.} & \multicolumn{2}{c}{Oracle} \\
          \cmidrule(lr){2-5} \cmidrule(lr){7-8} 
          & 4 & 8 & 16 & 32 & & Pass@1 & Avg. Len \\ 
          \midrule
          LLaDA-Base      & 22.4 & 51.4 & 55.8 & 48.4 & 44.5 & \textbf{87.6} & 9.5 \\
          DiffuCoder-Base & 24.1 & 51.5 & 59.8 & 47.6 & 45.8 & \textbf{87.4} & 8.6 \\
          DreamCoder-Base & 24.0 & 55.6 & 63.2 & 54.3 & 47.0 & \textbf{88.1} & 8.6 \\
          \bottomrule
        \end{tabular}
      \end{sc}
    \end{small}
  \end{center}
  \vskip -0.1in
\end{table*}

\textbf{Infilling Task Formulation.} The infilling task \cite{humaneval_infilling, fried2022incoder, text_infilling} aims to generate a semantically consistent middle segment $M$ of length $L$, given a prefix context $P$ and a suffix context $S$. Formally, the complete sequence is $\mathbf{x} = [P; M; S]$. While this formulation applies broadly, a critical constraint in diffusion-based infilling is that the length $L$ must be pre-specified to initialize the masked input state $\mathbf{x}^T$. Specifically, the model constructs an initial noisy input $\mathbf{x}^T = [P; \text{\texttt{[MASK]}}^L; S]$ and denoises the masked region conditioned on the bidirectional context. Since the optimal length $L^*$ is typically unknown a priori, a mismatched $L$ inevitably degrades generation quality, as the model cannot dynamically adjust the span size.

\textbf{First-Step Denoising Confidence.} To address the unknown length challenge without auxiliary training, we leverage the intrinsic uncertainty estimates provided by the diffusion model. We focus specifically on the first-step denoising, which corresponds to the model's initial prediction $p_\theta(\mathbf{x}^0 \mid \mathbf{x}^T)$ given the fully masked input state $\mathbf{x}^T = [P; \text{\texttt{[MASK]}}^L; S]$. This prediction represents the model's ``first guess'' of the complete sequence based on the bidirectional context. Crucially, we isolate the probabilities corresponding to the masked segment $M$. Let $p_j(v)$ denote the probability assigned to token $v$ at the $j$-th position of the generated sequence. We define the average first-step confidence $\Phi(L)$ over the masked indices $\mathcal{I}_{mask}$ as:
\begin{equation}
    \Phi(L) = \frac{1}{L} \sum_{j \in \mathcal{I}_{mask}} \max_{v \in \mathcal{V}} p_j(v),
\end{equation}
where $\mathcal{I}_{mask}$ represents the set of indices corresponding to the middle segment $M$, and $\mathcal{V}$ is the vocabulary. This metric serves as a computationally efficient proxy for semantic compatibility, allowing us to probe for the optimal length before committing to the full denoising trajectory.

\section{Analysis: Length-Confidence Dynamics} 
\label{sec:analysis}

\textbf{Sensitivity of Diffusion Infilling to Mask Length.} Existing DLMs \cite{llada, ye2025dream} typically follow a training pipeline of pre-training followed by instruction fine-tuning (SFT). During pre-training, the model learns to reconstruct text under random masks, while during SFT, it is optimized for instruction-response pairs. While effective for standard chat generation, this paradigm lacks explicit supervisory signals for \texttt{<eos>} or \texttt{<padding>} tokens within a masked span. Consequently, these models inherently lack the ability to automatically adjust infilling lengths. As illustrated in \Cref{fig:infilling_example}, generation quality highly depends on the preset [MASK] length: an underestimated length forces semantic truncation, while an overestimated length causes the model to hallucinate redundant content to fill the remaining slots. Experiments confirm that providing the Oracle length (the token count of the ground-truth answer) significantly boosts performance (see \Cref{tab:pass1_analysis}), suggesting that length discovery is the primary bottleneck for unleashing the potential of diffusion models.

\begin{figure*}[ht]
  \vskip 0.1in
  \begin{center}
    \centerline{\includegraphics[width=2.1\columnwidth]{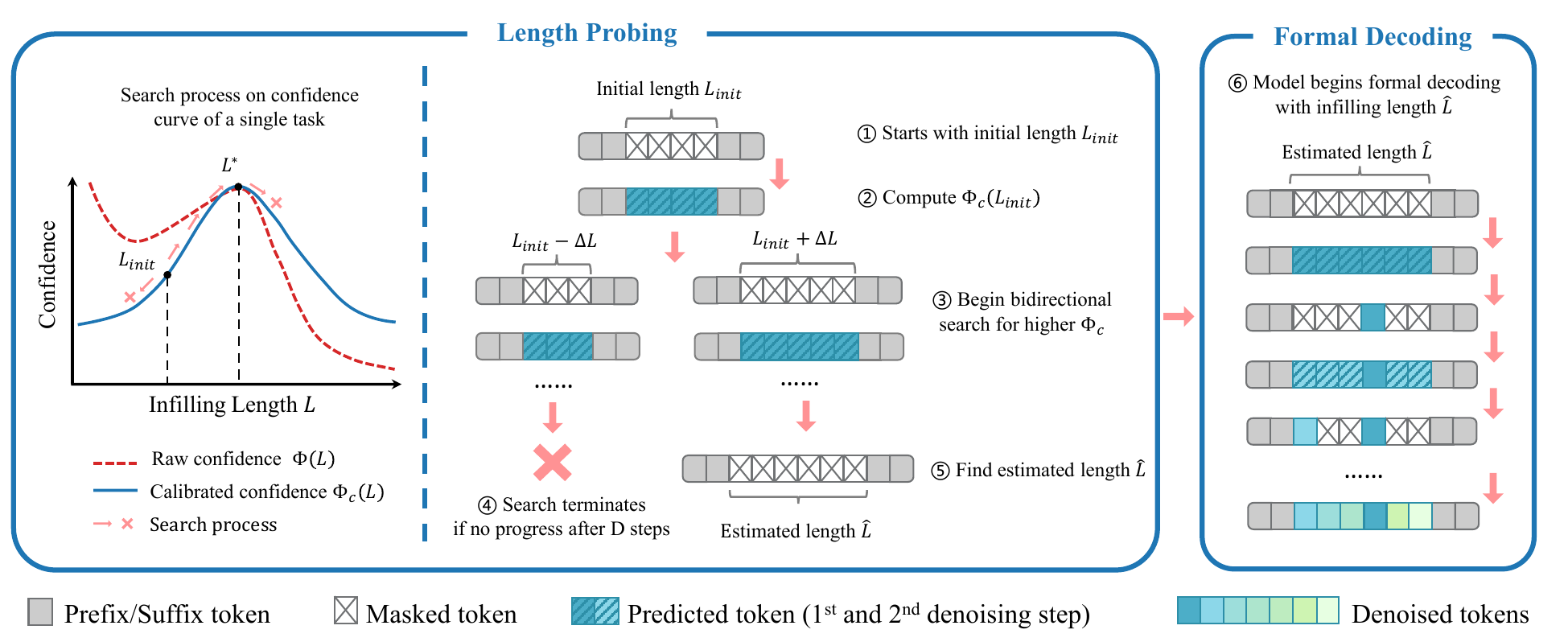}}
    \caption{\textbf{Calibrated Adaptive Length Framework.} The method operates in two stages: (1) \textbf{Length Probing}, where a bidirectional hill-climbing search identifies an estimated length $\hat{L}$ to approximate the Oracle length $L^*$ by maximizing the calibrated first-step confidence $\Phi_c(L)$; and (2) \textbf{Formal Decoding}, where the model performs standard iterative denoising initialized with the discovered length $\hat{L}$.}
    \label{fig:method}
  \end{center}
  \vskip -0.2in
\end{figure*}

\textbf{Oracle Peak: The Inherent Signal of Semantic Completeness.} Although diffusion models lack an explicit stopping mechanism, we discover that a strong signal of logical completeness is encoded within their probability distributions. Inspired by confidence-guided adaptive length methods~\cite{daedal}, we systematically analyze the evolution of the average first-step denoising confidence $\Phi(L - L^*)$ as a function of the infilling length $L$ and the Oracle length $L^*$. We observe that, under fixed contextual constraints, the confidence curve exhibits a significant and stable local maximum near the Oracle length (see \Cref{fig:oracle_peak}), which we term the \textit{Oracle Peak}. This phenomenon reveals that diffusion models can discriminate semantic completeness at the very onset of denoising: when the allocated mask space matches the logical ground truth in terms of information entropy, the model's prediction certainty reaches a maximum. This provides a solid physical basis for extracting length information without auxiliary training.

\textbf{Quantifying Systematic Length Bias.} While observing the Oracle Peak, we also identify a systematic phenomenon that interferes with the average first-step denoising confidence $\Phi(L)$, which we term \textit{Length Bias}. Specifically, $\Phi(L)$ exhibits an overall trend of rapid initial decline followed by gradual stabilization as $L$ increases, often causing the Oracle Peak to be obscured by background noise. This bias inherently reflects the monotonic decay of the model's prediction certainty as the search space expands under zero-information input (full mask). As shown by the red line in \Cref{fig:oracle_peak}, the raw $\Phi(L - L^*)$ values on the left side of the Oracle length are significantly higher than those on the right, displaying clear asymmetry. This semantics-agnostic bias constitutes the primary obstacle to cross-length comparison and needs to be eliminated through calibration.

\textbf{Calibrating Length Bias via Fitting.} To address the multi-scale decay of Length Bias, we fit a double-exponential function $B(L) = a e^{-bL} + c e^{-dL} + e$ using 100 samples from the HumanEval-Infilling benchmark with LLaDA-8B-Base (see \Cref{fig:length_bias}). Crucially, to isolate structural bias, we exclude data points near the ground-truth length during fitting, and these samples are strictly excluded from subsequent evaluations. We define the calibrated confidence $\Phi_c(L)$ as the ratio of raw confidence to the fitted bias:
\begin{equation}
    \Phi_c(L) = \frac{\Phi(L)}{B(L)}.
\end{equation}
Mechanistically, $B(L)$ represents the expected confidence under a zero-information prior governed solely by structural constraints. Thus, $\Phi_c(L)$ functions as a signal-to-noise ratio (SNR) metric, effectively decoupling semantic signals from length-dependent statistical bias. As shown by the blue curve in \Cref{fig:oracle_peak}, $\Phi_c(L-L^*)$ exhibits excellent symmetry around the Oracle length, rendering the Oracle Peak a globally prominent optimal solution. This confirms that, once structural noise is removed, the DLM's inherent confidence becomes a reliable indicator for length discovery. Specific fitting details are provided in Appendix~\ref{app:bias_fitting}. We employ this single fitted $B(L)$ across all subsequent experiments, demonstrating its cross-model and cross-task robustness in \Cref{sec:experiment}.

\section{Method}
\label{sec:method}

Building on the statistical findings in \Cref{sec:analysis}, we propose CAL (Calibrated Adaptive Length), a training-free framework for infilling length discovery. As illustrated in \Cref{fig:method}, CAL introduces a length probing stage prior to DLM formal decoding. This stage formulates length selection as a discrete optimization problem over a calibrated confidence landscape, aiming to approximate the Oracle length to enhance infilling performance.

\textbf{Optimization Objective.} As demonstrated in \Cref{sec:analysis}, DLMs achieve optimal infilling performance when provided with the Oracle length, a condition often signaled by an Oracle Peak in the first-step denoising confidence. Therefore, our primary goal is to find an estimated length $\hat{L}$ that best approximates the Oracle length $L^*$ by leveraging this intrinsic signal. However, directly maximizing the raw confidence $\Phi(L)$ is suboptimal because the Oracle Peak is typically obscured by systematic Length Bias. To address this limitation, we formulate length discovery as an optimization problem maximizing the calibrated confidence $\Phi_c(L)$. Consequently, we determine the Oracle length estimate $\hat{L}$ via the following maximization:
\begin{equation} 
    \hat{L} = \arg\max_{L \in \mathcal{R}} \Phi_c(L),
\end{equation}
where $\mathcal{R}$ denotes the search space of candidate lengths.

\textbf{Bidirectional Search Algorithm.} Exhaustively computing $\Phi_c(L)$ for every candidate length is computationally inefficient. However, our analysis in \Cref{sec:analysis} suggests that the Oracle Peak exhibits a quasi-unimodal property. Based on this observation, we employ a bidirectional hill-climbing strategy to locate the optimum efficiently. As detailed in \Cref{alg:length_discovery}, the algorithm starts from an initial heuristic estimate $L_{init}$ and probes the local trend of $\Phi_c(L)$ in both increasing and decreasing directions. Although noise may exist, the overall trend facilitates rapid convergence. The search terminates when the calibrated confidence fails to improve for $D$ consecutive steps. Following length discovery, we proceed to formal decoding by executing the full denoising process with the discovered length $\hat{L}$. Since the discovery stage requires only single-pass forward computations, it improves infilling performance with acceptable additional latency. A detailed case study illustrating this process and its robustness is provided in Appendix~\ref{app:case_studies}.

\begin{algorithm}[tb]
  \caption{Adaptive Length Discovery}
  \label{alg:length_discovery}
  \begin{algorithmic}
    \STATE {\bfseries Input:} Context $P, S$, Model $\mathcal{M}$, Bias $B(L)$
    \STATE {\bfseries Hyperparams:} Step $\Delta L$, Tolerance $D$, Init $L_{init}$, Max $L_{max}$
    \STATE $\hat{L} \gets L_{init}, \quad \hat{\Phi}_c \gets \Phi(L_{init}) / B(L_{init})$
    
    \FOR{$dir \in \{+1, -1\}$}
      \STATE $L \gets L_{init} + dir \cdot \Delta L, \quad cnt \gets 0$
      \WHILE{$L \in [1, L_{max}] \land cnt < D$}
        \STATE $\Phi_c \gets \Phi(L) / B(L)$
        \IF{$\Phi_c > \hat{\Phi}_c$}
          \STATE $\hat{L}, \hat{\Phi}_c, cnt \gets L, \Phi_c, 0$ \COMMENT{Update best length}
        \ELSE
          \STATE $cnt \gets cnt + 1$
        \ENDIF
        \STATE $L \gets L + dir \cdot \Delta L$
      \ENDWHILE
    \ENDFOR
    
    \STATE {\bfseries return} $\text{Denoise}(\mathcal{M}, P, S, \hat{L})$
  \end{algorithmic}
\end{algorithm}
\section{Experiment}
\label{sec:experiment}

\subsection{Experimental Setup}
\label{sec:experimental_setup}

We evaluate the effectiveness of CAL on both code and text infilling benchmarks.

\textbf{Code Infilling.} We utilize the HumanEval-Infilling benchmark \cite{humaneval_infilling}, a standard dataset for Python code completion, evaluating on both Single-Line and Multi-Line tasks. Crucially, the 100 samples used for Length Bias fitting in \cref{sec:analysis} were strictly excluded from the test set. We apply our method to four representative discrete diffusion language models: the general-purpose LLaDA-8B (Base and Instruct) \cite{llada}, and the code-specialized DiffuCoder-7B-Base \cite{diffucoder} and DreamCoder-Base-7B \cite{dreamcoder}. Additionally, we adapt DAEDAL \cite{daedal}, a confidence-based training-free method originally designed for chat tasks, as a baseline on LLaDA-8B-Base. All models employ greedy sampling for formal decoding, and we report Pass@1 as the primary metric.

\textbf{Text Infilling.} To assess generalization across diverse domains, we conduct experiments on three datasets: ROCStories (short five-sentence stories) \cite{rocstories}, CS Abstracts (academic writing) \cite{enable_infilling}, and Yelp Reviews (informal user-generated content) \cite{yelp}. For each sample, we randomly mask a contiguous span of 2 to 8 tokens. Experiments are conducted using LLaDA-8B-Base. We employ greedy sampling and report BLEU-2 and ROUGE-L scores. For detailed hyperparameters and implementation specifics, please refer to Appendix~\ref{app:experimental_setup_details}.

\begin{table*}[t]
  \caption{\textbf{Code Infilling results.} We compare the Pass@1 scores of various models under four fixed-length settings ($L=4, 8, 16, 32$) against our proposed adaptive length discovery algorithm (+ ours). Our method consistently outperforms fixed-length baselines across all initial length configurations. Stps. denotes the average number of search steps  incurred during the length discovery phase.}
  \label{tab:code_infilling}
  \begin{center}
    \begin{small}
      \begin{sc}
        \setlength{\tabcolsep}{4pt} 
        \begin{tabular}{lcccccccc | cc}
          \toprule
          \multirow{2}{*}{Model} & \multicolumn{2}{c}{Length = 4} & \multicolumn{2}{c}{Length = 8} & \multicolumn{2}{c}{Length = 16} & \multicolumn{2}{c}{Length = 32} & \multicolumn{2}{|c}{Avg.} \\
          \cmidrule(lr){2-3} \cmidrule(lr){4-5} \cmidrule(lr){6-7} \cmidrule(lr){8-9} \cmidrule(lr){10-11}
          & Pass@1 & Stps. & Pass@1 & Stps. & Pass@1 & Stps. & Pass@1 & Stps. & Pass@1 & Stps. \\ 
          \midrule
          \multicolumn{11}{c}{Single-Line Infilling} \\
          \midrule
          LLaDA-Base        & 22.7 & -- & 51.0 & -- & 56.6 & -- & 49.0 & -- & 44.8 & --  \\
          \quad + daedal   & 29.9& -- & 52.8& -- & 53.4& -- & 43.1& -- & 44.8& --  \\
          \quad + cal      & \textbf{70.4} & 11.6 & \textbf{73.6} & 12.3 & \textbf{65.2} & 14.8 & \textbf{52.6} & 15.7 & \textbf{65.5} & 13.6  \\
          LLaDA-Instruct    & 23.0 & -- & 54.9 & -- & 64.0 & -- & 57.0 & -- & 49.7 & --  \\
          \quad + cal      & \textbf{74.6} & 12.2 & \textbf{76.9} & 13.0 & \textbf{69.4} & 14.3 & \textbf{58.7} & 15.3 & \textbf{69.9} & 13.7  \\
          DiffuCoder-Base   & 24.5& -- & 51.5 & -- & 60.1& -- & 47.7& -- & 46.0& --  \\
          \quad + cal      & \textbf{73.1}& 11.1& \textbf{74.8}& 12.4& \textbf{68.4}& 14.6& \textbf{55.8}& 18.2& \textbf{68.0}& 14.1  \\
          DreamCoder-Base   & 24.7& -- & 55.4& -- & 63.1& -- & 53.9& -- & 49.3& --  \\
          \quad + cal      & \textbf{75.1}& 11.1& \textbf{76.2}& 12.4& \textbf{70.5}& 14.6& \textbf{59.0}& 18.1& \textbf{70.2}& 14.1\\
          \midrule
          \multicolumn{11}{c}{Multi-Line Infilling} \\
          \midrule
          LLaDA-Base        & 4.6  & -- & 10.8 & -- & 18.7 & -- & 27.6 & -- & 15.4 & --  \\
          \quad + daedal   & 6.0 & -- & 12.3&  -- & 20.4 & -- & 26.5 & -- & 16.3 & --  \\
          \quad + cal      & \textbf{17.1} & 12.0 & \textbf{21.2} & 13.3 & \textbf{27.1} & 14.7 & \textbf{31.7} & 15.3 & \textbf{24.3} & 13.8 \\
          LLaDA-Instruct    & 4.7 & -- & 11.7 & -- & 21.4 & -- & 30.4 & -- & 17.0 & --  \\
          \quad + cal      & \textbf{18.8} & 12.0 & \textbf{22.8} & 13.1 & \textbf{30.2} & 13.8 & \textbf{34.0} & 14.8 & \textbf{26.5} & 13.4  \\
          DiffuCoder-Base   & 5.0& -- & 11.7& -- & 21.6& -- & 30.8& -- & 17.3& --  \\
          \quad + cal      & \textbf{19.4}& 12.1& \textbf{24.5}& 13.4& \textbf{32.0}& 14.3& \textbf{38.0}& 16.6& \textbf{28.5}& 14.1\\
          DreamCoder-Base   & 5.1& -- & 12.4& -- & 22.9& -- & 33.1& -- & 18.4& --  \\
          \quad + cal      & \textbf{19.9}& 12.1& \textbf{24.4}& 13.3& \textbf{31.9}& 14.5& \textbf{37.3}& 16.9& \textbf{28.4}& 14.2  \\
          \bottomrule
        \end{tabular}
      \end{sc}
    \end{small}
  \end{center}
  \vskip -0.1in
\end{table*}

\subsection{Code Infilling Results}
Table \ref{tab:code_infilling} presents the comprehensive results on code infilling. 

\textbf{Consistent Improvement over Baselines.} Our adaptive length discovery method consistently achieves superior performance across all models and initial length configurations compared to fixed-length baselines. For instance, on the Single-Line task, our approach improves the average Pass@1 of LLaDA-Base from 44.8\% to 65.5\%. As indicated by the \textit{Stps.} column, achieving this substantial performance gain requires only 11 to 18 additional first-step forward passes on average. This overhead is acceptable given the substantial capability improvements.

\textbf{Outperforming Local-Confidence Methods.} Notably, our method significantly outperforms DAEDAL, a confidence-based adaptive strategy originally designed for DLM chat tasks. While effective for open-ended generation, DAEDAL's strategy of locally inserting masks based on token-level confidence proves suboptimal for infilling. As shown in Table \ref{tab:code_infilling}, it occasionally underperforms even the fixed-length baseline, particularly when the initial length is overestimated (e.g., $L=16$ or $32$). This performance gap highlights two key advantages of our method. First, our \textit{bidirectional search} allows for both increasing and decreasing the span length, whereas DAEDAL is primarily limited to extending it. Second, relying on \textit{global average confidence} proves more effective than local heuristics. Since DLMs model the joint probability of all tokens holistically, the average confidence over the entire span serves as a more reliable indicator of semantic completeness.

\textbf{Performance Trends and Robustness.} We observe that the performance gain diminishes as the initial length increases. We attribute this to two factors: (1) \textit{Signal Decay.} The HumanEval-Infilling dataset is biased towards short completions (average Oracle length $\approx 9.5$, see \Cref{tab:pass1_analysis}). Consequently, a large initial length (e.g., $L=32$) implies a significant deviation from the ground truth, causing the Oracle Peak signal to decay and become obscured by background noise (as evidenced by confidence fluctuations when $|L - L^*| > 10$ in \Cref{fig:oracle_peak}). (2) \textit{Increased Prediction Difficulty.} As the masked span expands, the first-step reconstruction task becomes inherently more challenging, reducing the model's capacity to produce a sharp confidence peak. Yet, our method consistently outperforms fixed-length baselines even in these difficult scenarios, proving its robustness.

\begin{table*}[t]
  \caption{\textbf{Text Infilling results.} Comparisons of BLEU-2 and ROUGE-L scores on three text datasets using LLaDA-8B-Base. The results are averaged over all test samples. Fixed-length baselines are reported for preset lengths of 2, 4, and 8 tokens. Our adaptive method consistently outperforms the baselines, demonstrating robust generalization across different textual domains.}
  \label{tab:text_infilling}
  \begin{center}
    \begin{small}
      \begin{sc}
        \setlength{\tabcolsep}{4.5pt} 
        \begin{tabular}{lccccccc | cc}
          \toprule
          \multirow{2}{*}{Dataset} & \multirow{2}{*}{Model} & \multicolumn{2}{c}{Length = 2} & \multicolumn{2}{c}{Length = 4} & \multicolumn{2}{c}{Length = 8} & \multicolumn{2}{|c}{Avg.} \\
          \cmidrule(lr){3-4} \cmidrule(lr){5-6} \cmidrule(lr){7-8} \cmidrule(lr){9-10}
          & & BLEU-2 & ROUGE-L & BLEU-2 & ROUGE-L & BLEU-2 & ROUGE-L & BLEU-2 & ROUGE-L \\ 
          \midrule
          \multirow{2}{*}{Stories} & LLaDA-Base     & 9.1 & 25.8 & 18.0 & 37.9 & 17.8 & 37.6 & 15.0 & 33.8  \\
                                      & \quad + cal  & \textbf{17.6}& \textbf{35.7}& \textbf{19.8}& \textbf{38.4}& \textbf{21.2} & \textbf{40.7}& \textbf{19.5}& \textbf{38.3}\\
          \midrule
          \multirow{2}{*}{Abstract}   & LLaDA-Base     & 8.7 & 26.8 & 17.8 & 38.7 & 19.3 & 39.5 & 15.3 & 35.0  \\
                                      & \quad + cal  & \textbf{17.1}& \textbf{36.5}& \textbf{19.6}& \textbf{39.6}& \textbf{22.1}& \textbf{42.4}& \textbf{19.6}& \textbf{39.5}\\
          \midrule
          \multirow{2}{*}{Yelp} & LLaDA-Base     & 5.9 & 20.1 & 11.4 & \textbf{30.5} & 12.5 & 30.3 & 9.9 & 27.0  \\
                                      & \quad + cal  & \textbf{10.4}& \textbf{27.8}& \textbf{11.9}& 30.2& \textbf{13.6}& \textbf{32.5}& \textbf{12.0}& \textbf{30.2}\\
          \bottomrule
        \end{tabular}
      \end{sc}
    \end{small}
  \end{center}
  \vskip -0.1in
\end{table*}

\begin{table*}[t]
\caption{\textbf{Ablation Study of Length Bias (LB) Calibration.} We demonstrate the necessity and robustness of LB calibration across three dimensions: (a) varying initial lengths on code infilling using LLaDA-Base; (b) different model on code infilling ($L_{init}=8$), where LLaDA, Diffu., and Dream. correspond to LLaDA-Instruct, DiffuCoder-Base, and DreamCoder-Base, respectively; and (c) varying initial lengths on text infilling. In all settings, our calibrated search (\textit{w/ LB}) significantly outperforms both the fixed-length baseline and the uncalibrated search (\textit{w/o LB}).}
  \label{tab:LB_ablation}
  \begin{center}
    \begin{small}
      \begin{sc}
        \begin{tabular}{l ccc | ccc | ccc}
          \toprule
          \multirow{3}{*}{Method} & \multicolumn{3}{c|}{(a) Length (Code)} & \multicolumn{3}{c|}{(b) Models (Code, $L=8$)} & \multicolumn{3}{c}{(c) Length (Text)} \\
          \cmidrule(lr){2-4} \cmidrule(lr){5-7} \cmidrule(lr){8-10}
           & \multicolumn{3}{c|}{Pass@1} & \multicolumn{3}{c|}{Pass@1} & \multicolumn{3}{c}{BLEU-2 / ROUGE-L} \\
           & $L=4$ & $L=8$ & $L=16$ & LLaDA & Diffu. & Dream. & $L=2$ & $L=4$ & $L=8$ \\
          \midrule
          Fixed Length          & 22.7 & 51.0 & 56.6 & 54.9 & 51.5 & 55.4& 9.1 / 25.8 & 18.0 / 37.9 & 17.8 / 37.6 \\
          \midrule
          CAL \textit{w/o LB} & 35.8& 59.6 & 60.8& 66.5 & 69.1& 67.1 & 14.5 / 32.1& 16.9 / 35.2& 19.5 / 38.4\\
          CAL \textit{ w/ LB} & \textbf{70.4} & \textbf{73.6} & \textbf{65.2} & \textbf{76.8} & \textbf{74.8}& \textbf{76.2}& \textbf{17.6} / \textbf{35.7}& \textbf{19.8} / \textbf{38.4}& \textbf{21.2} / \textbf{40.7}\\
          \midrule
          \textit{LB Improvement} & \textit{+34.6}& \textit{+14.0}& \textit{+4.4}& \textit{+10.3}& \textit{+5.7}& \textit{+9.1}& \textit{+3.1} / \textit{+3.6}& \textit{+2.9} / \textit{+3.2}& \textit{+1.7} / \textit{+2.3}\\
          \bottomrule
        \end{tabular}
      \end{sc}
    \end{small}
  \end{center}
  \vskip -0.1in
\end{table*}

\subsection{Text Infilling Results}
Table \ref{tab:text_infilling} extends our evaluation to general text domains, including short stories, scientific abstracts, and reviews. 

\textbf{Generalization Across Domains.} 
Our method consistently surpasses fixed-length baselines in both BLEU-2 and ROUGE-L metrics across all datasets. For instance, on the Stories dataset, we achieve a relative improvement of 30.0\% in BLEU-2 (from 15.0 to 19.5) compared to the average baseline performance. These results indicate that the discovered length signal is not limited to code syntax but reflects a fundamental semantic property of diffusion language models, proving effective for diverse natural language generation tasks.

\textbf{Modality-Specific Performance Variations.} 
We observe that the performance gains in text infilling are more modest compared to code infilling. We attribute this discrepancy to the inherent differences between the two modalities. Code is governed by strict syntax and rigid logical constraints (e.g., variable definitions), which significantly narrow the search space for the ground truth. This constraint allows the model to produce a sharp and distinct Oracle Peak. In contrast, natural language is inherently ambiguous and open-ended; the rich semantic space allows for multiple valid completions of varying lengths. This ambiguity leads to a more diffuse probability distribution, resulting in a less prominent Oracle Peak signal compared to the code domain.

\subsection{Ablation Study}

\textbf{Necessity of Length Bias Calibration.} 
To validate the critical role of bias calibration, we perform comprehensive comparisons across different initial lengths, models, and data modalities (see Table \ref{tab:LB_ablation}). Specifically, we utilize the HumanEval-Infilling Single-Line task for code infilling and the ROCStories dataset for text infilling. The results indicate that while the uncalibrated CAL method (w/o LB) generally outperforms the fixed-length baseline, applying Length Bias calibration yields significantly more pronounced improvements.

\textbf{Impact at short lengths.}
Calibration is particularly vital for short sequences. As shown in panels (a) and (c) of Table \ref{tab:LB_ablation}, the performance gain is substantially larger when the initial length is small. This trend holds for both code and text tasks. The reason lies in the nature of Length Bias: as shown in Figure \ref{fig:length_bias}, the bias curve decays most steeply at shorter lengths. Consequently, without calibration, raw confidence scores are artificially high for extremely short sequences. This often misleads the search into selecting truncated solutions. Our calibration effectively corrects this skew, allowing the model to recover the true Oracle signal.

\textbf{Effective Across Models and Domains.}
We fit the calibration function $B(L)$ using only LLaDA-Base on a subset of HumanEval-Infilling data. Yet, this same function demonstrates excellent performance on other models (DiffuCoder, DreamCoder) and even distinct modalities (text infilling). This success across models and domains suggests that Length Bias is an inherent statistical property of DLMs, rather than a model-specific artifact. Consequently, a single fitted bias function can be robustly transferred to various diffusion models,  eliminating the need for re-fitting on each new model.
\section{Related Work}
\label{sec:related_work}

\textbf{Diffusion Language Models.} Discrete diffusion models have recently emerged as a compelling alternative to auto-regressive models, enabling probabilistic modeling of token sequences through iterative denoising. Foundational works \cite{discrete_diffusion, hoogeboom2021argmax} introduced discrete corruption processes, which were further refined by improved training objectives \cite{zheng2023reparameterized, lou2023discrete}. Within this domain, Masked Diffusion Models (MDMs) have become a dominant paradigm, utilizing masking kernels to achieve scalable training \cite{sahoo2024simple, shi2024simplified}. Recent efforts have pushed these models to the billion-parameter frontier: LLaDA \cite{llada} pioneered large-scale training from scratch, while its successor LLaDA-1.5 \cite{llada_1.5} has integrated reinforcement learning for alignment, and LLaDA-V \cite{llada_v} has extended to multimodal understanding. Parallel research, such as DiffuLLaMA \cite{gong2024scaling} and Dream \cite{ye2025dream}, explored adapting pre-trained AR models into diffusion frameworks. Furthermore, specialized models like DiffuCoder \cite{diffucoder} and DreamCoder \cite{dreamcoder} have been developed specifically for code generation. While leveraging the bidirectional nature of diffusion, most prior work has focused primarily on standard instruction-following capabilities.

\textbf{Generative Infilling.} 
Early approaches to generative infilling predominantly relied on encoder-decoder architectures, such as BART \cite{lewis2020bart}, T5 \cite{raffel2020exploring}, and Insertion Transformers \cite{pmlr-v97-stern19a}, which treat infilling as a span corruption and reconstruction task. With the dominance of decoder-only models, research shifted toward adapting AR architectures for infilling. Techniques like Fill-In-the-Middle (FIM) \cite{humaneval_infilling, fried2022incoder} and blank-infilling objectives \cite{enable_infilling, glm} were introduced to enable bidirectional context awareness within a causal generation framework. More recently, the rise of diffusion language models \cite{llada, gong2024scaling, ye2025dream} has offered a new paradigm. Thanks to their non-causal nature, these models naturally model the joint distribution of the prefix, middle, and suffix. While standard DLMs typically require fixed lengths, specialized variants such as DDOT \cite{ddot}, FlexMDM \cite{anyorder}, and others \cite{dreamon, controllable_generation} have been proposed to enable variable-length generation capabilities.

\textbf{Adaptive Length Control in DLMs.} 
Standard diffusion language models typically handle length variations in instruction-following tasks by initializing an oversized canvas and filling unused positions with \texttt{<eos>} or padding tokens \cite{llada, gong2024scaling, ye2025dream}. While recent works like Rainbow Padding \cite{rainbow_padding} have optimized this padding mechanism, these strategies are primarily tailored for chat-style interactions rather than infilling. To enable flexible-length generation for infilling, several training-based approaches have been proposed. For instance, \citet{ddot} achieved flexible-length infilling via optimal transport, while \citet{edit_flow} and \citet{anyorder} abandoned the fixed-length mask-decode paradigm in favor of operations like insertion and deletion. Similarly, \citet{dreamon} introduced special tokens \texttt{<expand>} and \texttt{<delete>} to dynamically adjust sequence length, while LaViDa \cite{li2025lavida} relied on a specialized training stage with FIM objectives to achieve variable-length infilling. However, these methods necessitate extensive pre-training or task-specific fine-tuning.

In contrast, training-free approaches such as DAEDAL \cite{daedal} attempted to dynamically adjust length at inference time in chat tasks by monitoring local token confidence. Similarly, \citet{controllable_generation} prompted DLMs to generate \texttt{<null>} tokens as placeholders to regulate length in controllable JSON generation. However, these methods rely on explicit termination signals or rigid structural constraints, which are often absent in general  infilling scenarios. Our work bridges this gap by proposing a training-free framework that leverages the global first-step confidence of the entire masked region. By calibrating the systematic Length Bias, we approximate the optimal infilling length without relying on additional training.
\section{Conclusion}
\label{sec:conclusion}

In this paper, we address a critical limitation of Diffusion Language Models (DLMs) in infilling tasks: their sensitivity to pre-specified mask lengths. Our analysis reveals that DLMs naturally encode a length signal within their first-step denoising confidence. We characterize this through two key phenomena: the \textit{Oracle Peak}, which signals semantic completeness, and \textit{Length Bias}, a systematic confidence decay that must be calibrated. Leveraging these insights, we propose CAL (Calibrated Adaptive Length), a training-free framework for adaptive length discovery. By calibrating the bias and employing an efficient bidirectional search, CAL approximates the optimal infilling length prior to formal decoding. Extensive experiments on code and text benchmarks demonstrate that our approach consistently outperforms fixed-length baselines across diverse models and domains.

\textbf{Limitations and Future Work.} 
Our framework currently relies on a pre-fitted bias function. While our analysis confirms its robustness across domains, developing an online estimation method that adapts dynamically to each input represents a promising direction for future improvement. Additionally, although the search overhead is acceptable given the significant performance gains, the introduced latency might still be a consideration for ultra-low-latency real-time applications. Furthermore, our current evaluation focuses on single-span infilling; extending the calibrated search mechanism to handle multi-span or disjoint infilling scenarios remains a valuable direction for future research.




\section*{Impact Statement}
This paper presents work whose goal is to advance the field of Machine Learning, specifically focusing on the controllability of diffusion language models for infilling tasks. There are many potential societal consequences of our work, none of which we feel must be specifically highlighted here.

\bibliography{my_paper}
\bibliographystyle{icml2026}

\newpage
\appendix
\onecolumn
\section{Length Bias Fitting Details}
\label{app:bias_fitting}

\textbf{Data Sampling and Oracle Exclusion Strategy.} We randomly sampled 100 single-line infilling tasks with varying ground-truth lengths from the HumanEval-Infilling benchmark \cite{humaneval_infilling}. Crucially, these tasks were strictly excluded from the test set used in our main experiments to prevent any data leakage. For each task $i$, we fixed the prefix $P_i$ and suffix $S_i$, probing the first-step denoising confidence across a wide range of lengths $\mathcal{L}_{test} = \{1, 2, 4, 6, 12, 16, 24, 32, 48, 64, 96, 128\}$. To decouple the systematic Length Bias from semantic signals, we implemented a strict \textit{Oracle exclusion strategy}: any probe length $L$ falling within the neighborhood of the Oracle length $L^*_i$ (specifically, $L \in [L^*_i - 4, L^*_i + 4]$) was discarded. This ensures that the fitted curve $B(L)$ captures the background entropy decay rather than the model's semantic confidence.

\textbf{Double-Exponential Fitting.} Based on the characteristic trend of \textit{rapid initial decay followed by stabilization} (visualized in \Cref{fig:length_bias}), we adopted a double-exponential model $B(L) = a e^{-bL} + c e^{-dL} + e$. The first exponential term accounts for the sharp dilution of local constraints in short sequences, while the second term models the long-tail asymptotic behavior. We solved for the parameters using non-linear least squares, weighting each length point $L$ by $w_L = 1/\sqrt{N_L}$ to account for sample size variations.

\textbf{Fitted Parameters and Robustness.} For the LLaDA-8B-Base model \cite{llada}, the optimal parameters were determined as: $a=1.00, b=1.77$ (fast decay), $c=0.56, d=0.06$ (slow decay), and $e=0.24$ (baseline). Although fitting on different subsets may yield minor parameter fluctuations, our ablation studies (see \Cref{tab:LB_ablation}) demonstrate that the derived bias function is highly robust. It generalizes effectively across diverse models (DiffuCoder \cite{diffucoder}, DreamCoder \cite{dreamcoder}) and domains (text infilling \cite{rocstories, enable_infilling}), confirming that Length Bias is a universal statistical property of diffusion masking. We provide the full fitting script in our supplementary material to ensure reproducibility.

\section{Experimental Setup Details}
\label{app:experimental_setup_details}

\textbf{Hyperparameters and Resources.} For our adaptive length discovery method (CAL), we set the search step size $\Delta L = 1$ for all tasks to ensure fine-grained length probing. The search tolerance is set to $D = 4$ for code infilling tasks and $D = 2$ for text infilling tasks. Regarding the initial length configuration, we evaluated four settings for code infilling ($L_{init} \in \{4, 8, 16, 32\}$) and three settings for text infilling ($L_{init} \in \{2, 4, 8\}$). All experiments were conducted on a node equipped with 8 NVIDIA A40 (40GB) GPUs. 

\textbf{Input Format.} We adopted the standard infilling paradigm where the input sequence is constructed as [Prefix] [MASK] [Suffix]. During the denoising process, the tokens in the prefix and suffix regions remain clamped (i.e., unmasked and fixed), while the model performs length adjustment and decoding exclusively on the masked span.

\textbf{Baseline DAEDAL Implementation.} We adapted DAEDAL \cite{daedal}, a training-free dynamic length method originally for chat tasks, to serve as our infilling baseline. The original DAEDAL algorithm consists of two stages: (1) an initial length extension based on \texttt{<eos>} confidence, and (2) a runtime expansion based on low-confidence tokens during decoding. For the infilling adaptation, we omitted the first stage because DLMs in infilling mode typically do not generate explicit \texttt{<eos>} tokens, rendering the stop-signal-based extension ineffective. We focused on the second stage, where the span length is dynamically increased if tokens exhibit low confidence. Based on tuning results (see \Cref{tab:daedal_ablation}), we adjusted the confidence thresholds to better suit the infilling distribution: the low-confidence threshold was raised to $0.3$ (compared to $0.1$ in the original paper), as we found $0.1$ to be too conservative for infilling scenarios. The high-confidence threshold was maintained at the original value of $0.9$. The expansion span was set to $2$ tokens to facilitate fine-grained length adjustment. The maximum generation length was constrained to $64$ for Single-Line tasks and $128$ for Multi-Line tasks.

\section{Parameter Ablation}

\subsection{DAEDAL Threshold Sensitivity}
We first examine the impact of the low-confidence threshold in the DAEDAL baseline. As shown in \Cref{tab:daedal_ablation}, the performance of DAEDAL is highly sensitive to this threshold. A lower threshold (e.g., 0.1 or 0.2) makes the model conservative in expanding the length, failing to correct underestimated lengths (e.g., $L=4$). Conversely, a higher threshold (e.g., 0.5) encourages aggressive expansion, which benefits short initial lengths but degrades performance when the initial length is already sufficient ($L=16$ or $32$) by introducing unnecessary redundancy. We selected a threshold of 0.4 as the optimal trade-off, achieving the highest average Pass@1 (44.8\%) across all settings, which matches the fixed-length baseline on average but provides better robustness for shorter initializations.

\label{app:daedal_parameter_ablation}
\begin{table}[h]
  \caption{\textbf{Impact of DAEDAL low-confidence threshold on code infilling performance.} Evaluations on HumanEval-Infilling (SingleLine) by LLaDA-8B-Base.}
  \label{tab:daedal_ablation}
  \begin{center}
    \begin{small}
      \begin{sc}
        \setlength{\tabcolsep}{8pt} 
        \begin{tabular}{c cccc c}
          \toprule
          \multirow{2}{*}{Threshold} & \multicolumn{4}{c}{Initial Length (Pass@1)} & \multirow{2}{*}{Avg.} \\
          \cmidrule(lr){2-5}
           & $L=4$ & $L=8$ & $L=16$ & $L=32$ & \\
          \midrule
          Fixed Length  & 22.7  & 51.0 & \textbf{56.6} & \textbf{49.0}  & \textbf{44.8}  \\
          \midrule
          0.1   & 22.7 & 49.8 & 54.8 & 46.0 & 43.3 \\
          0.2   & 23.6 & 50.0 & 54.8 & 43.5 & 43.0 \\
          0.3   & 26.4 & 51.2 & 55.6 & 43.5 & 44.2 \\
          0.4   & 29.9 & 52.8 & 53.4 & 43.1 & \textbf{44.8} \\
          0.5   & \textbf{32.9} & \textbf{54.2} & 50.0 & 38.8 & 44.0 \\
          \bottomrule
        \end{tabular}
      \end{sc}
    \end{small}
  \end{center}
\end{table}

\subsection{CAL Search Parameters}
We further investigate the robustness of our CAL method with respect to its two key search parameters: the tolerance $D$ (stopping criterion) and the step $\Delta L$ (search granularity).

\textbf{Impact on Code Infilling.} \Cref{tab:cal_code_ablation} presents the results on the HumanEval-Infilling benchmark. We observe a clear trade-off between search accuracy and computational cost. Increasing the tolerance $D$ from 2 to 4 leads to a performance gain (from 71.4\% to 74.4\%), as it prevents the search from terminating prematurely at local optima. However, this comes at the cost of increased search steps. Beyond $D=4$, the performance plateaus while the cost continues to rise. Regarding step size, a finer granularity ($\Delta L=1$) outperforms a coarser one ($\Delta L=2$), confirming that precise length estimation is critical for code generation. Consequently, we adopt $D=4$ and $\Delta L=1$ as the default configuration for code tasks.

\textbf{Impact on Text Infilling.} For text infilling tasks, as shown in \Cref{tab:parameter_ablation}, the method exhibits greater robustness to parameter variations. Even with a minimal tolerance of $D=1$, CAL significantly outperforms the fixed-length baseline. Increasing $D$ to 2 or 3 yields marginal improvements in BLEU-2 and ROUGE-L scores. This suggests that the confidence landscape in natural language is smoother than in code, making the Oracle Peak easier to locate. To balance performance and efficiency, we select $D=2$ and $\Delta L=1$ for all text infilling experiments.

\label{app:code_parameter_ablation}
\begin{table}[h]
  \caption{\textbf{Impact of CAL search parameters on code infilling.} Evaluations on HumanEval-Infilling (SingleLine) by LLaDA-8B-Base with $L_{init}=8$. Stps. denotes the average number of search steps  incurred during the length discovery phase.}
  \label{tab:cal_code_ablation}
  \begin{center}
    \begin{small}
      \begin{sc}
        \setlength{\tabcolsep}{10pt} 
        \begin{tabular}{cc cc}
          \toprule
          Tolerance $D$ & Step $\Delta L$ & Pass@1 & Stps. \\
          \midrule
          \multicolumn{2}{c}{Fixed Length ($L=8$)} & 51.4 & -- \\
          \midrule
          2 & 1 & 71.4 & 7.3 \\
          2 & 2 & 65.9 & 6.4 \\
          3 & 1 & 73.1 & 9.8 \\
          4 & 1 & \textbf{74.4} & 12.2 \\
          5 & 1 & \textbf{74.4} & 14.4 \\
          \bottomrule
        \end{tabular}
      \end{sc}
    \end{small}
  \end{center}
\end{table}

\label{app:text parameter ablation}
\begin{table}
  \caption{\textbf{Impact of CAL parameters on text infilling performance.} Evaluations are conducted on ROCStories dataset by LLaDA-8B-Base.}
  \label{tab:parameter_ablation}
  \begin{center}
    \begin{small}
    \begin{sc}
        \setlength{\tabcolsep}{4pt} 
        \begin{tabular}{cccccccc | cc}
          \toprule
          \multirow{2}{*}{Tolerance $D$} & \multirow{2}{*}{Step $\Delta L$} & \multicolumn{2}{c}{Length = 2} & \multicolumn{2}{c}{Length = 4} & \multicolumn{2}{c}{Length = 8} & \multicolumn{2}{|c}{Avg.} \\
          \cmidrule(lr){3-4} \cmidrule(lr){5-6} \cmidrule(lr){7-8} \cmidrule(lr){9-10}
          & & BLEU-2 & ROUGE-L & BLEU-2 & ROUGE-L & BLEU-2 & ROUGE-L & BLEU-2 & ROUGE-L \\ 
          \midrule
          \multicolumn{2}{c}{Fixed Length} & 9.1 & 25.8 & 18.0 & 37.9 & 17.8 & 37.6 & 15.0 & 33.8  \\
          \midrule
          1 & 1 & 15.0 & 32.6 & 20.2 & 39.5 & 21.0 & 40.9 & 18.7 & 37.7  \\
          1 & 2 & 18.0 & 36.4 & 19.5 & 38.3 & 20.7 & 40.2 & 19.4 & \textbf{38.3}  \\
          2 & 1 & 17.6 & 35.7 & 19.8 & 38.4 & 21.2 & 40.7 & 19.5& \textbf{38.3}\\
          3 & 1 & 18.8 & 37.1 & 19.5 & 37.9 & 20.8 & 39.9 & \textbf{19.7} & \textbf{38.3}  \\
          \bottomrule
        \end{tabular}
      \end{sc}
    \end{small}
  \end{center}
  \vskip -0.1in
\end{table}

\section{Case Studies}
\label{app:case_studies}

To provide a concrete understanding of how CAL operates, we present a detailed case study on the task HumanEval/0/L3~\cite{humaneval_infilling}. As shown in \Cref{tab:task_structure}, this task requires the model to infill the core logic for calculating the distance between two numbers within a nested loop. The ground-truth middle segment is \texttt{distance = abs(elem - elem2)\textbackslash n}, which consists of exactly 10 tokens using the LLaDA tokenizer.

\textbf{Search Process Analysis.}
\Cref{tab:case_studies_combined} demonstrates the robustness of our bidirectional search. Whether starting from an underestimated length ($L=4$, Table a) or an overestimated one ($L=16$, Table b), the calibrated confidence $\Phi_c(L)$ effectively suppresses length bias and guides the algorithm to find at the Oracle length ($L=10$).

\textbf{Visualization of the Oracle Peak.}
\Cref{fig:method} plots the confidence landscape for this task. It clearly shows that while the raw confidence (red) decays due to structural noise (blue dashed), the calibrated score (blue solid) reveals a prominent Oracle Peak at $L=10$. Although local fluctuations exist, the calibrated confidence metric $\Phi_c(L)$ reaches its maximum at the Oracle length.

\begin{table}[h]
\centering
\caption{\textbf{Illustration of an Infilling Task Structure.} This example (HumanEval/0/L3) demonstrates the decomposition of a code generation task into a prefix ($P$), a ground-truth middle ($M^*$), and a suffix ($S$). The model's goal is to recover the core semantic logic within the masked span.}
\begin{minipage}{0.7\columnwidth}
\label{tab:task_structure}
\centering
\begin{small}
    \begin{tabularx}{\textwidth}{X}
    \toprule
    \textbf{Prefix ($P$)} \\ 
    \midrule
    \texttt{from typing import List} \\
    \texttt{} \\
    \texttt{def has\_close\_elements(numbers, threshold):} \\
    \texttt{~~""" Check if in given list of numbers, are any} \\
    \texttt{~~two numbers closer to each other than given} \\
    \texttt{~~threshold.} \\
    \texttt{~~>>> has\_close\_elements([1.0, 2.0, 3.0], 0.5)} \\
    \texttt{~~False} \\
    \texttt{~~>>> has\_close\_elements([1.0, 2.8, 3.0, 4.0, 5.0, 2.0], 0.3)} \\
    \texttt{~~True} \\
    \texttt{~~"""} \\
    \texttt{~~for idx, elem in enumerate(numbers):} \\
    \texttt{~~~~for idx2, elem2 in enumerate(numbers):} \\
    \texttt{~~~~~~if idx != idx2:}
    \\ 
    \midrule
    \textbf{Middle ($M^*$)} \\ \midrule
    \colorbox{gray!30}{\makebox[\dimexpr\linewidth-2\tabcolsep][l]{\texttt{~~~~~~~~distance = abs(elem - elem2)}}} \\ 
    \midrule
    \textbf{Suffix ($S$)} \\ 
    \midrule
    \texttt{~~~~~~if distance < threshold:} \\
    \texttt{~~~~~~~~return True} \\
    \texttt{} \\
    \texttt{~~return False}
    \\ 
    \bottomrule
    \end{tabularx}
\end{small}
\end{minipage}
\vskip -0.1in
\end{table}

\begin{table*}[t]
\caption{\textbf{Case Studies of Bidirectional Length Probing.} We illustrate the search process for Task HumanEval/0/L3 on LLaDA-8B-Instruct (matching the example in \Cref{fig:infilling_example}) starting from different initial lengths. $\Phi(L)$ denotes the raw confidence, $B(L)$ the fitted bias, and $\Phi_c(L)$ the calibrated confidence. In both scenarios, the algorithm accurately identifies the Oracle length $L^*=10$ (highlighted), where the calibrated confidence $\Phi_c(L)$ is maximized and the model recovers the ground-truth logic.}
\label{tab:case_studies_combined}

\begin{subtable}{\textwidth}
    \centering
    \caption{Starting from an underestimated length ($L_{init}=4$).}
    \begin{small}
    \setlength{\tabcolsep}{4.5pt}
    \begin{tabularx}{\textwidth}{l ccc X}
    \toprule
    \textbf{Probe} $L$ & $\Phi(L)$ & $B(L)$ & $\Phi_c(L)$ & \textbf{First-Step Denoising Prediction} \\
    \midrule
    $L=4$ & 0.956 & 0.681 & 1.403 & \texttt{distance =\textbackslash n} \\
    $L=5$ & 0.941 & 0.655 & 1.437 & \texttt{distance = elem\textbackslash n} \\
    $L=6$ & 0.926 & 0.631 & 1.468 & \texttt{distance = elem2\textbackslash n} \\
    $L=7$ & 0.820 & 0.608 & 1.348 & \texttt{distance = abs(elem2\textbackslash n} \\
    $L=8$ & 0.979 & 0.587 & 1.669 & \texttt{distance = elem - elem2\textbackslash n} \\
    $L=9$ & 0.790 & 0.566 & 1.395 & \texttt{distance = abs(elem elem2)\textbackslash n} \\
    \rowcolor{gray!30} $L=10$ & 0.997 & 0.547 & \textbf{1.821} & \texttt{distance = abs(elem - elem2)\textbackslash n} \\
    $L=11$ & 0.886 & 0.529 & 1.674 & \texttt{distance = abs(elem - elem2)\textbackslash n\textbackslash n} \\
    $L=12$ & 0.604 & 0.513 & 1.179 & \texttt{distance = abs(elem abs -2 elemabs)\textbackslash n} \\
    $L=13$ & 0.583 & 0.497 & 1.173 & \texttt{distance = abs(elem - abs(elem - elem2)\textbackslash n} \\
    $L=14$ & 0.472 & 0.482 & 0.979 & \texttt{distance = abs \ \ \ \ \ \ distance abs abs(elem - elem2)\textbackslash n} \\
    \midrule
    $L=3$ & 0.887 & 0.713 & 1.244 & \texttt{distance\textbackslash n} \\
    $L=2$ & 0.926 & 0.766 & 1.209 & \texttt{\textbackslash n} \\
    $L=1$ & 1.000 & 0.938 & 1.066 & \texttt{\textbackslash n} \\
    \bottomrule
    \end{tabularx}
    \end{small}
\end{subtable}

\vspace{0.2in} 

\begin{subtable}{\textwidth}
    \centering
    \caption{Starting from an overestimated length ($L_{init}=16$).}
    \begin{small}
    \setlength{\tabcolsep}{4.5pt}
    \begin{tabularx}{\textwidth}{l ccc X}
    \toprule
    \textbf{Probe} $L$ & $\Phi(L)$ & $B(L)$ & $\Phi_c(L)$ & \textbf{First-Step Denoising Prediction } \\
    \midrule
    $L=16$ & 0.524 & 0.454 & 1.153 & \texttt{distance = abs(elem - elem2) abs(elem - elem2)\textbackslash n} \\
    $L=17$ & 0.520 & 0.442 & 1.177 & \texttt{distance = abs(elem - elem2)\textbackslash n abs(elem - elem2)\textbackslash n} \\
    $L=18$ & 0.414 & 0.430 & 0.963 & \texttt{distance = abs(elem - elem2) distance abs abs(elem - elem2)\textbackslash n} \\
    $L=19$ & 0.425 & 0.419 & 1.014 & \texttt{distance = abs(elem - elem2\textbackslash n distance = abs(elem - elem2)\textbackslash n} \\
    $L=20$ & 0.389 & 0.409 & 0.951 & \texttt{distance = abs(elem - elem2\textbackslash n\textbackslash n distance = abs(elem - elem2)\textbackslash n} \\
    $L=21$ & 0.351 & 0.399 & 0.879 & \texttt{distance = abs(elem - elem2\textbackslash n\textbackslash n distance =(elem abs(elem - elem2)\textbackslash n} \\
    \midrule
    $L=15$ & 0.427 & 0.468 & 0.913 & \texttt{distance = abs(elem - -2) abs - elem2)\textbackslash n} \\
    $L=14$ & 0.472 & 0.482 & 0.979 & \texttt{distance = abs \ \ \ \ \ \ distance abs abs(elem - elem2)\textbackslash n} \\
    $L=13$ & 0.583 & 0.497 & 1.173 & \texttt{distance = abs(elem - abs(elem - elem2)\textbackslash n} \\
    $L=12$ & 0.604 & 0.513 & 1.179 & \texttt{distance = abs(elem abs -2 elemabs)\textbackslash n} \\
    $L=11$ & 0.886 & 0.529 & 1.674 & \texttt{distance = abs(elem - elem2)\textbackslash n\textbackslash n} \\
    \rowcolor{gray!30} $L=10$ & 0.997 & 0.547 & \textbf{1.821} & \texttt{distance = abs(elem - elem2)\textbackslash n} \\
    $L=9$ & 0.790 & 0.566 & 1.395 & \texttt{distance = abs(elem elem2)\textbackslash n} \\
    $L=8$ & 0.979 & 0.587 & 1.669 & \texttt{distance = elem - elem2\textbackslash n} \\
    $L=7$ & 0.820 & 0.608 & 1.348 & \texttt{distance = abs(elem2\textbackslash n} \\
    $L=6$ & 0.926 & 0.631 & 1.468 & \texttt{distance = elem2\textbackslash n} \\
    \bottomrule
    \end{tabularx}
    \end{small}
\end{subtable}

\vskip -0.1in
\end{table*}

\begin{figure*}[ht]
  \vskip 0.1in
  \begin{center}
    \centerline{\includegraphics[width=0.6\columnwidth]{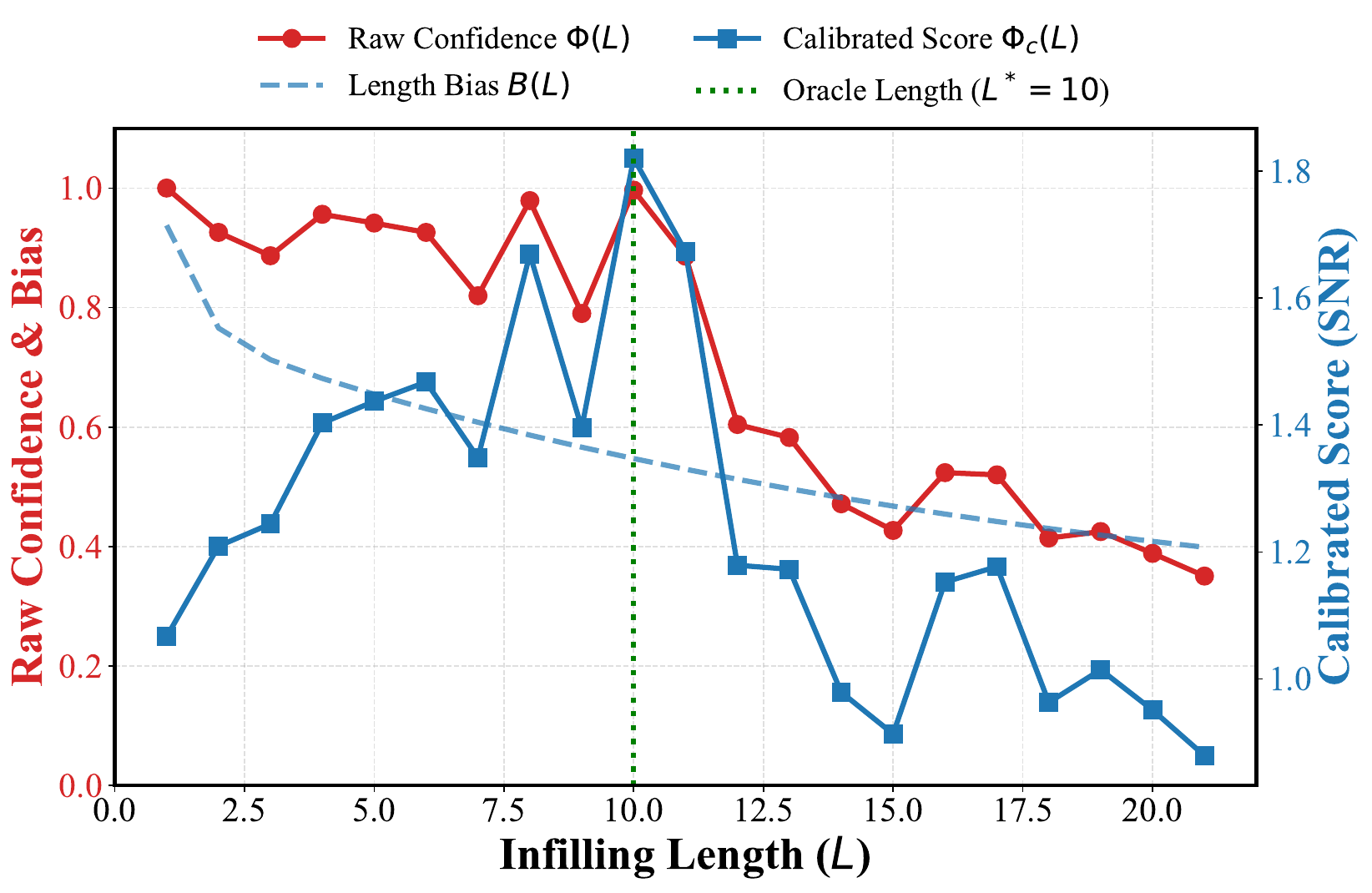}}
    \caption{\textbf{Dynamics of the confidence landscape during bidirectional search.} We compare the raw confidence $\Phi(L)$ (red) and the calibrated score $\Phi_c(L)$ (blue solid). While the raw signal is biased towards shorter lengths, our calibration against the systematic bias $B(L)$ (blue dashed) reveals a prominent Oracle Peak at the ground-truth length $L^*=10$ (green dotted line).}
    \label{fig:method}
  \end{center}
  \vskip -0.2in
\end{figure*}


\end{document}